\documentclass[12pt]{article}
\usepackage{chicagor}
\newtheorem{THEOREM}{Theorem}[section]
\newenvironment{theorem}{\begin{THEOREM} \hspace{-.85em} {\bf :} }%
                        {\end{THEOREM}}
\newtheorem{LEMMA}[THEOREM]{Lemma}
\newenvironment{lemma}{\begin{LEMMA} \hspace{-.85em} {\bf :} }%
                      {\end{LEMMA}}
\newtheorem{COROLLARY}[THEOREM]{Corollary}
\newenvironment{corollary}{\begin{COROLLARY} \hspace{-.85em} {\bf :} }%
                          {\end{COROLLARY}}
\newtheorem{PROPOSITION}[THEOREM]{Proposition}
\newenvironment{proposition}{\begin{PROPOSITION} \hspace{-.85em} {\bf :} }%
                            {\end{PROPOSITION}}
\newtheorem{DEFINITION}[THEOREM]{Definition}
\newenvironment{definition}{\begin{DEFINITION} \hspace{-.85em} {\bf :} \rm}%
                            {\end{DEFINITION}}
\newtheorem{CLAIM}[THEOREM]{Claim}
\newenvironment{claim}{\begin{CLAIM} \hspace{-.85em} {\bf :} \rm}%
                            {\end{CLAIM}}
\newtheorem{EXAMPLE}[THEOREM]{Example}
\newenvironment{example}{\begin{EXAMPLE} \hspace{-.85em} {\bf :} \rm}%
                            {\end{EXAMPLE}}
\newtheorem{REMARK}[THEOREM]{Remark}
\newenvironment{remark}{\begin{REMARK} \hspace{-.85em} {\bf :} \rm}%
                            {\end{REMARK}}
\newcommand{\thm}{\begin{theorem}}
\newcommand{\lem}{\begin{lemma}}
\newcommand{\pro}{\begin{proposition}}
\newcommand{\dfn}{\begin{definition}}
\newcommand{\rem}{\begin{remark}}
\newcommand{\xam}{\begin{example}}
\newcommand{\cor}{\begin{corollary}}
\newcommand{\ethm}{\end{theorem}}
\newcommand{\elem}{\end{lemma}}
\newcommand{\epro}{\end{proposition}}
\newcommand{\edfn}{\bbox\end{definition}}
\newcommand{\erem}{\bbox\end{remark}}
\newcommand{\exam}{\bbox\end{example}}
\newcommand{\ecor}{\end{corollary}}
\newcommand{\beqn}{\begin{equation}}
\newcommand{\eeqn}{\end{equation}}
\newcommand{\bbox}{\vrule height7pt width4pt depth1pt}

\newcommand{\clm}{\begin{claim}}
\newcommand{\eclm}{\end{claim}}
\newcommand{\union}{\cup}
\newcommand{\inter}{\cap}
\newcommand{\IR}{\mbox{$I\!\!R$}}

\renewcommand{\phi}{\varphi}
\renewcommand{\P}{{\cal P}}

\newcommand{\U}{{\cal U}}

\newcommand{\ie}{i.e.,~}

\newcommand{\ol}{\setlength{\itemsep}{0pt}\begin{enumerate}}
\newcommand{\eol}{\end{enumerate}\setlength{\itemsep}{-\parsep}}
\newcommand{\ul}{\setlength{\itemsep}{0pt}\begin{itemize}}
\newcommand{\dl}{\setlength{\itemsep}{0pt}\begin{description}}
\newcommand{\edl}{\end{description}\setlength{\itemsep}{-\parsep}}
\newcommand{\eul}{\end{itemize}\setlength{\itemsep}{-\parsep}}

\newcommand{\commentout}[1]{}

\newcommand{\bi}{\begin{itemize}}
\newcommand{\ei}{\end{itemize}}
\newcommand{\be}{\begin{enumerate}}
\newcommand{\ee}{\end{enumerate}}

\newcommand{\red}{\mbox{\em red}}
\newcommand{\blue}{\mbox{\em blue}}

\newcommand{\yellow}{\mbox{\em yellow}}

\newcommand{\Bel}{{\rm Bel}}
\newcommand{\Plaus}{{\rm Plaus}}
\newcommand{\lives}{\mbox{\em lives}}
\newcommand{\lE}{\underline{E}}
\newcommand{\uE}{\overline{E}}
\newcommand{\sfa}{{\bf a}}

\newcommand{\intensionp}[1]{[\![ #1 ]\!]}

\usepackage{url}

\begin{document}

\title{Using Sets of Probability Measures to Represent Uncertainty%
\thanks{The material in this chapter is taken, often verbatim, from 
[Halpern 2003], which the
reader is encouraged to consult for further details and references.}}

\author{Joseph Y. Halpern%
\thanks{Supported in part by NSF under grants  
CTC-0208535, ITR-0325453, and IIS-0534064, by ONR under grants
N00014-00-1-03-41 and 
N00014-01-10-511, by the DoD Multidisciplinary University Research
Initiative (MURI) program administered by the ONR under
grants N00014-01-1-0795 and N00014-04-1-0725, and by AFOSR under grant
F49620-02-1-0101.}\\
Cornell University\\
Ithaca, NY 14853\\
halpern@cs.cornell.edu\\
http://www.cs.cornell.edu/home/halpern
}

\maketitle

\section{Introduction}

\nocite{Hal31}
Despite its widespread acceptance, there are some problems in using 
probability to represent uncertainty.
Perhaps the most serious is that probability is not
good at representing ignorance.  The following two examples illustrate
the problem.

\xam\label{coinxam} Suppose that a coin is tossed once.
There are two possible worlds, $h$ and
$t,$ corresponding to the two possible outcomes.
If the coin is known to be fair,
it seems reasonable to assign probability $1/2$ to each of these worlds.
However, suppose that the coin has an unknown bias
(where the \emph{bias} of a coin is the probability that it lands heads.)
How should this be
represented?  One approach might be to continue to take heads and tails
as the elementary outcomes and,  applying the principle of indifference,
assign them both probability $1/2$, just as in the case of a fair coin.
However, there seems to be a significant qualitative
difference between a fair coin and a coin of unknown bias.  
This difference has some pragmatic consequences.  For example, as Kyburg
(e.g., in \cite{Kyb1}) has pointed out, the
assumption that heads and tails have probability $1/2$, together with
the assumption that consecutive coin tosses are independent implies
that, if the coin is tossed 1,000,000 times, then the probability 
that the coin will land heads somewhere between 498,000 and 502,000
times is  greater than .999.  This certainly doesn't seem something that
an agent who has no idea of the bias of the coin should know!
\exam

\xam\label{marbles} Suppose that a bag contains 100 marbles; 30 are
known to be red, and the remainder are known to be either blue or yellow,
although the exact proportion of blue and yellow is not known.
What is the likelihood that a marble taken out of the bag is yellow?
This can be modeled with three possible worlds, $\red,$ $\blue,$ and
$\yellow,$ one for each of the possible
outcomes.  It seems reasonable to assign probability .3 to the outcome
to choosing a red marble, and thus probability .7 to choosing either
blue or yellow, but what probability should be assigned to the other
two outcomes?

Empirically, it is clear that people do {\em not\/} use probability to
represent the uncertainty in this example.
For example, consider the following three bets.  In each case a marble is
chosen from the bag. 
\begin{itemize}
\item $B_r$ pays \$1 if the marble is red, and 0 otherwise;
\item $B_b$ pays \$1 if the marble is blue, and 0 otherwise;
\item $B_y$ pays \$1 if the marble is yellow, and 0 otherwise.
\end{itemize}
People invariably prefer $B_r$ to both $B_b$ and $B_y,$ and they are
indifferent between $B_b$ and $B_y$.  The fact that they are indifferent
between $B_b$ ad $B_y$ suggests that they view it equally likely that
the marble chosen is blue and that it is yellow.  This seems reasonable;
the problem statement provides no reason to prefer blue to yellow, or
vice versa.  However, if the probability of drawing a red marble is
taken to be .3, then
the probability of drawing a blue marble and that of drawing a yellow marble
are both .35, which suggests that $B_y$ and $B_b$ should both be
preferred to $B_r$.  

Moreover, now consider the following three bets: 
\begin{itemize}
\item $B_{ry}$ pays \$1 if the marble is red or yellow, and 0 otherwise;
\item $B_{by}$ pays \$1 if the marble is blue or yellow, and 0 otherwise.
\end{itemize}
While most people prefer $B_r$ to $B_b$, most also prefer $B_{by}$ to
$B_{ry}$.  There is no probability measure on $\{b,r,y\}$ that
would both make $b$ more likely than $r$ and make $\{b,r\}$ less likely
than $\{b,y\}$.
(This is essentially Ellsberg's \citeyear{Ellsberg61}
paradox; I return to this issue in Section~\ref{sec:decision}.)
\exam

One natural way of representing uncertainty in both of these cases is by
using a \emph{set} of probability measures, rather than a single
measure.  For example, the uncertainty in  Example~\ref{coinxam} can be
represented by the set $\P_m = \{\mu_a: a \in
[0,1]\}$ of probability measures on $\{h,t\}$, where $\mu_a$
gives $h$ probability $a$.  In Example~\ref{marbles},
the uncertainty 
can be represented using the set $\P_u = \{\mu'_a: a \in [0,.7]\}$ of
probability measures on $\{\red,\blue,\yellow\}$, 
where $\mu'_a$ gives $\red$ probability 
.3, $\blue$ probability $a,$ and $\yellow$ probability $.7-a$.

In the rest of this paper, I explore the use of sets of probability
measures as a representation of uncertainty.

\section{Lower and Upper Probability and Dutch Book Arguments}
Let $\P$ be a set of probability measures all defined on 
all subsets of a finite set $W$ of possible worlds.%
\footnote{The assumptions that $W$ is finite and that every subset of $W$
is \emph{measurable}, that is, in the domain of every probability
measure $\mu \in \P$, are made for ease of exposition only.
They can both easily be dropped.}
Given a set $X$ of real numbers, let $\sup X,$ the {\em
supremum\/}\index{supremum\/} (or just {\em sup\/}) of 
$X,$ be the {\em least upper bound of $X$}---the smallest real number
that is at least as large as all the elements in $X$.  That is, $\sup X =
\alpha$
if $x \le \alpha$ for all $x \in X$ and if, for all $\alpha' < \alpha,$
there is some $x \in X$ such that $x > \alpha'$.  
For example, if $X = \{1/2, 3/4, 7/8, 15/16, \ldots\}$, then 
$\sup X = 1$.  Similarly, $\inf X,$
the {\em infimum\/} (or just {\em inf\/}) of $X$\index{infimum (inf)},
is the greatest lower bound of $X$---the 
largest real number that is less than or equal to every element in $X$.
For $U \subseteq W,$ define
$$\begin{array}{l}
\P(U) = \{\mu(U): \mu \in \P\},\\
\P_*(U) = \inf \P(U), \mbox{ and }\\
\P^*(U) = \sup \P(U).
\end{array}$$
$\P_*(U)$\glossary{\glospaalow} is called the {\em lower
probability\/}\index{lower/upper probability|(} of $U,$ and $\P^*(U)$ is
called the {\em upper probability\/} of $U$.\glossary{\glospaaup}
If $\P^*(U) = \P_*(U)$ for all subsets $U$ of $W$, then it is easy to
see that $\P$ must be a singleton $\{\mu\}$, and $\P^* = \P_* = \mu$.  
In general, of course, $\P^* \ne \P_*$.
For a set $U$, the difference $\P^*(U) - \P_*(U)$ can be viewed as
characterizing our ignorance about $U$.  In Example~\ref{marbles}, there
is uncertainty about the likelihood of $\red$ being chosen, but there is
no ignorance: the likelihood is exactly $.3$.  This is captured by
$\P_2$: $(\P_2)_*(\red) = (\P_2)^*(\red) = .3$.  On the other hand,
there is ignorance about the likelihood of $\blue$ and $\yellow$ being
chosen.  And, indeed, $(\P_2)_*(\blue) = 0$ and
$(\P_2)^*(\blue) = .7$, and similarly for $\yellow$.

While lower and upper probabilities seem natural, how reasonable is it
to use them to represent uncertainty?  I investigate this question in a
number of different contexts in the next few sections.  For now, I
briefly consider one of the most prominent justifications for
probability, the \emph{Dutch book argument}, which goes back to 
Ramsey \citeyear{Ram} and de Finetti 
\citeyear{DeFinetti31,DeFinetti37}, 
and see how it fares in the
context of sets of probabilities.

Roughly speaking, the Dutch book argument says that if odds do not
act like probabilities, then there is a collection of bets that
guarantees a sure loss.  Somewhat more precisely,
suppose that an agent must post
odds for each subset of a set $W$.  If the agent chooses odds of,
say, 4:5 on 
$U \subseteq W$, then this is supposed to mean that the agent is
willing to accept a bet of any size for or against $U$.  If a bookie
bets $\$k$ on $U$, then if $U$ happens (i.e., if the actual world is in
$U$---it is assumed 
that this can always be determined), then the bookie wins $\$9/4k$; if
not, the bookie loses the $\$k$.  Similarly, if the bookie bets $\$k$
against $U$, then if the $U$ happens, the bookie loses the $\$k$, and if
not, then the bookie wins $\$9/5k$.  In general, if the odds for $U$ are
$o_1: o_2$, then if a bookie  bets $\$k$ on $U$, then he wins $(o_1 +
o_2)/o_1$ if $U$ happens and loses the $\$k$ otherwise, and if he bets
against $U$, he loses $\$k$ if $U$ happens and wins $(o_1 + o_2)/o_2$
otherwise.  If the odds on $U$ are $o_1:o_2$, let $p_U$ be
$o_1/(o_1 + o_2)$.  The key claim is that, unless the numbers $p_U$ act
like probabilities (and, in particular, $p_W = 1$ and $p_{U \union V} =
p_U + p_V$ if $U$ and $V$ are disjoint), then the agent is irrational:
there is a \emph{Dutch book}, a collection of bets which guarantee a
loss for the agent.
Conversely, if the $p_U$'s do act like probabilities, then there is no
Dutch book. 

Does this mean it is irrational to use other representations of
uncertainty, such as sets of probability measures?  Many problems have
been noted with Dutch book arguments (see, for example,
\cite[pp.~89--91]{HowUrb}, \cite{Hajek07}).  Of most relevance here is the implicit
assumption that an agent can or is willing to post \emph{fair odds},
that is, odds for which he is indifferent between a bet for and against
a subset $U$ of $W$.  In the stock market, bid and ask prices are not
necessarily equal.  Suppose that instead of posting fair odds, the agent
were only willing to post the analogue of bid and ask prices; odds for
which he is willing to take a bet on $U$ and (lower) odds at which he is
willing to take a bet against $U$.   In that case, arguments similar in
spirit to those used by de Finetti and Ramsey can be used to show that
the agent is rational iff his odds determine lower and upper
probabilities (see \cite{Smith,Williams76}).  The key to making these
arguments 
precise is a characterization of lower and upper probabilities, which is
the subject of the next section.

\section{Charaterizing Lower and Upper Probability}
A probability measure on $W$ is a function $\mu: 2^W \rightarrow [0,1]$ 
characterized by two well-known properties:
\begin{itemize}
\item[P1.] $\mu(W) = 1$.
\item[P2.] $\mu(U_1 \union U_2) = \mu(U_1) + \mu(U_2)$ if $U_1$ and
$U_2$ are disjoint subsets of $W$.
\end{itemize}
Every probability measure satisfies P1 and P2, and every function from
$\mu: 2^W \rightarrow [0,1]$ satisfying P1 and P2 is a probability
measure. Property P2 is known as \emph{(finite) additivity}; note that
the fact that $\mu(\emptyset) = 0$ follows easily from P2; P1 and P2
together imply that $\mu(\overline{U}) = 1 - \mu(U)$.

Are there similar properties characterizing lower and upper probabilities?
It is easy to see that P1 continues to hold for both lower and upper
probabilities.  P2 does not hold, but 
lower probability is {\em superadditive\/}\index{superadditivity} and
upper probability is {\em 
subadditive,\/}\index{subadditivity} so that for disjoint sets $U$ and $V,$
\begin{equation}\label{pch3.eq7}
\begin{array}{l}
\P_*(U \union V) \ge \P_*(U) +
\P_*(V), \mbox{ and }\\
\P^*(U \union V) \le \P^*(U) + \P^*(V).
\end{array}\end{equation}
In addition, the relationship between lower and upper probability is
defined by
\begin{equation}\label{pch3.eq7.5}
\P_*(U) = 1 - \P^*(\overline{U}).
\end{equation}
(I leave the straightforward proof of these results to the reader.)

While (\ref{pch3.eq7}) and (\ref{pch3.eq7.5}) hold for all lower and
upper probabilities, these properties do not completely characterize them.
For example, the following
property holds for lower and upper probabilities if $U$
and $V$ are disjoint:
\begin{equation}\label{pch3.eq8}
\P_*(U \union V) \le \P_*(U) + \P^*(V) \le \P^*(U \union V);
\end{equation}
moreover, (\ref{pch3.eq8}) does not follow from (\ref{pch3.eq7}) and
(\ref{pch3.eq7.5})
\cite{HalPuc00}. However, even adding (\ref{pch3.eq8}) to
(\ref{pch3.eq7}) and~(\ref{pch3.eq7.5}) does not provide a complete
characterization of lower and upper probabilities.
The property needed to get a complete characterization is somewhat
complex.  To state it precisely, say that 
a set $\U$ of subsets of $W$ {\em covers a subset $U$ 
of $W$ exactly $k$ times\/}\index{covers} if every element of $U$ is in
exactly $k$ sets 
in $\U$.  Consider the following property:
\begin{equation}\label{pch3.eq9}
\begin{array}{c}
\mbox{If $\U = \{U_1, \ldots, U_k\}$ covers $U$ exactly $m+n$ times and
covers $\overline{U}$}\\ 
\mbox{exactly $m$ times, then $\sum_{i=1}^k \P_*(U_i) \le m + n\P_*(U).$}
\end{array}
\end{equation}
(There is of course an analogous property for upper probability, with
$\le$ replaced by $\ge$.)
It is not hard to show that lower probabilities satisfy~(\ref{pch3.eq9})
and that (\ref{pch3.eq7}) and (\ref{pch3.eq8}) follow from
(\ref{pch3.eq9}) and (\ref{pch3.eq7.5}).
Indeed, in a precise sense, as Anger and Lembcke \citeyear{AL85}
show, 
(\ref{pch3.eq9}) completely characterizes lower probabilities (and
hence, together with (\ref{pch3.eq7.5}), upper probabilities as well).

\thm\label{thm:upchar} {\rm \cite{AL85}} Lower 
probability satisfies 
(\ref{pch3.eq9}). Conversely, if $f: 2^W \rightarrow [0,1]$ satisfies 
(\ref{pch3.eq9}) (with $\P_*$ replaced by $f$) and $f(W) = 1$, then there
exists a set $\P$ of probability measures such that $f = \P_*$.%
\footnote{Besides the characterization of Anger and Lembcke given in
Theorem~\ref{thm:upchar}, 
a number of other characterizations of lower and upper probability have
been given in the literature, all similar in spirit
\cite{r:giles82,r:huber76,Huber81,r:lorentz52,Williams76,Wolf77}.}
\ethm

Although I have been focusing on lower and upper probability, 
it is important to stress that sets of probability measures contain more 
information than is captured by their lower and upper probability, as
the following example shows.

\xam\label{xam:moreinfo}
Consider two variants of Example~\ref{marbles}.  In the first,
all that is known is that there are at most 50 yellow marbles and at most
50 blue marbles in a bag of 100 marbles; no information at all is given
about the number of red marbles.  In the second case, it is known that
there are exactly as many blue marbles as yellow marbles.  The first 
situation can be captured by the set $\P_3 = \{\mu: \mu(\blue) \le .5,
\mu(\yellow) \le .5\}$.  The second situation can be captured by the set $\P_4
= \{\mu: \mu(b) = \mu(y)\}$.  These sets of measures are obviously quite
different; in fact $\P_4$ is a strict subset of $\P_3$.  However, it is
easy to see that 
$(\P_3)_* = 
(\P_4)_*$ and, hence, that $\P_3^* = \P_4^*$.
Thus, the fact that blue and yellow have equal probability in every
measure in $\P_4$ has been lost by considering only lower and upper
probability.  I return to this issue 
in Section~\ref{sec:expectation}.
\exam

\section{Dempster-Shafer Belief Functions as Lower Probabilities}

The Dempster-Shafer theory of evidence, originally introduced by Arthur
Dempster \citeyear{Demp1,Demp} and then developed by Glenn Shafer
\citeyear{Shaf}, provides another approach 
to attaching likelihoods to events.  This approach starts out with a {\em
belief function}\index{belief function|(}
(sometimes called a {\em support function}).  Given a
set $W$ of possible worlds and $U \subseteq W,$
the belief in $U,$ denoted $\Bel(U),$\glossary{\glosbel} is a number in the
interval $[0,1]$.  A belief function $\Bel$
defined on a space $W$ must satisfy the following three properties:
\begin{itemize}
\index{B1--3|(}
\item[B1.] $\Bel(\emptyset) = 0$.
\item[B2.] $\Bel(W) = 1$.
\item[B3.]
$\Bel(\union_{i=1}^n U_i) \ge \sum_{i=1}^n \sum_{\{I \subseteq \{1, \ldots, n\}: |I| = i\}}
(-1)^{i+1} \Bel(\inter_{j \in I} U_j),$ for $n = 1, 2, 3, \ldots$.
\end{itemize}

B1 and B2 just say that, like probability measures, belief functions
follow the convention of using 0 and 1 to denote the minimum and maximum
likelihood.  B3 is closely related to the \emph{inclusion-exclusion}
rule for probability.  The inclusion-exclusion rule is used to compute
the probability of 
the union of (not necessarily disjoint) sets.
In the case of two sets $U$ and $V$, the rule says
$$
\mu(U \union V) =
\mu(U) + \mu(V) - \mu(U \inter V).
$$
In the case of three sets $U_1$, $U_2$, $U_3$, 
similar arguments show that
$$
\begin{array}{l}
\mu(U_1 \union U_2 \union U_3) = \\
\ \ \ \mu(U_1) + \mu(U_2) + \mu(U_3) - \mu(U_1 \inter U_2) -
\mu(U_1 \inter U_3) - \mu(U_2 \inter U_3) + \mu(U_1 \inter	U_2 \inter U_3).
\end{array}
$$
That is, the probability of the union of $U_1$, $U_2$, and $U_3$ can be
determined by adding the probability of the individual sets
(these are one-way intersections), subtracting the probability of the
two-way intersections, and adding the probability of the three-way
intersections.   The generalization of this rule to $k$ sets, with $=$
replaced by $\ge$, is just B3.  It follows that every probability
measure is a belief function.

If $U$ and $V$ are disjoint sets, then it easily follows from B1 and B3
that $\Bel(U \union V) \ge \Bel(U) + \Bel(V)$.  That is, $\Bel$ is
superadditive, just like a lower probability.  
And just like a lower probability,
$\Bel(U)$ can be viewed as providing a lower bound
on the likelihood of $U$.
Define $\Plaus(U) = 1 - \Bel(\overline{U})$.\glossary{\glosplaus}  $\Plaus$ is a {\em
plausibility function;\/}\index{plausibility function|(}
$\Plaus(U)$ is the
{\em plausibility\/} of $U$.   A plausibility
function bears the same relationship to a belief function that upper
probability bears to lower probability.  

By B2 and B3, for all subsets $U \subseteq W$, $1 = \Bel(W) \ge \Bel(U)
+ \Bel(\overline{U}),$ 
so $$\Plaus(U) = 1 - \Bel(\overline{U}) \ge \Bel(U).$$
Thus, for an event $U,$ the
interval $[\Bel(U),\Plaus(U)]$ can be viewed as describing the range of
possible values of the likelihood of $U$, just like $[\P_*(U), \P^*(U)]$.

There is in fact a deeper connection between belief functions and lower
probabilities: every belief function is a lower probability and the
corresponding plausibilty function is the corresponding upper
probability.

\thm\label{pch3.thm1} Given a belief function $\Bel$ defined on a space
$W,$ let $\P_{\Bel} =
\{\mu: \mu(U) \ge \Bel(U) \mbox{ for all } U \subseteq
W\}$.\glossary{\glospabel}  Then 
$\Bel = (\P_{\Bel})_*$ and $\Plaus = (\P_{\Bel})^*$.
\ethm

The converse of Theorem~\ref{pch3.thm1} does not follow, as the following
example shows.

\xam\label{xam:lowerprobnotbel} Suppose that $W = \{a,b,c,d\}$, $\P =
\{\mu_1, \mu_2\}$, $\mu_1(a) = \mu_1(b) = \mu_1(c) = \mu_1(d) = 1/4$,
and $\mu_2(a) = \mu_2(c) = 1/2$ (so that $\mu_2(b) = \mu_2(d) = 0$).
Let $U_1 = \{a,b\}$ and $U_2 = \{b,c\}$.  It is easy to check that
$\P_*(U_1) = \P_*(U_2) = 1/2$, $\P_*(U_1 \union U_2) = 3/4$, and
$\P_*(U_1 \inter U_2) = 0$.  $\P_*$ thus cannot be a belief function,
because it violates B3:
$$\P_*(U_1 \union U_2) < \P_*(U_1) + \P_*(U_2) - \P_*(U_1 \inter U_2).$$
\exam

Thus, lower probabilities are a strictly
more expressive representation of uncertainty than belief functions.

I remark that while belief functions can be understood (to
some extent) in terms of lower probability\index{lower/upper
probability}, this is not the only way of understanding them. 
Shafer, for example, views belief functions as a way of representing
\emph{evidence}; see \cite{HF2} for a discussion of these two ways of
understanding belief functions.

\section{Updating Sets of Probabilities}

Suppose that an agent's
uncertainty is defined in terms of a set $\P$ of probability
measures.  How should the agent update his beliefs in light of observing
an event $U$?  The obvious thing to do is to condition
each member of $\P$ on $U$.  This suggests that after observing $U,$ the
agent's uncertainty should be represented by the set
$\{\mu|U: \mu \in \P\}$ (where $\mu|U$ is the conditional probability
measure that results by conditioning $\mu$ on $U$).
There is one obvious issue that needs to be addressed:
What happens if $\mu(U) = 0$ for some $\mu \in \P$?  There are two
choices here:\ either to say that conditioning makes sense only if
$\mu(U) > 0$ for all $\mu \in
\P$ (\ie if $\P_*(U) > 0$) or to consider only those measures $\mu$ for
which $\mu(U) > 0$.  The latter choice is somewhat more general, so that
is what I use here.  Thus, I  define
$$\P|U = \{\mu|U: \mu \in \P, \, \mu(U) > 0\}.$$
Once the agent has a set $\P|U$ of conditional probability measures, it
is possible to consider lower and upper conditional probabilities.
However, note that the lower and upper conditional probabilities are not
determined by the lower and upper probabilities, as the following
example shows.  

\xam\label{lpnotcp} 
Let $\P_3$ and $\P_4$ be the sets of probability measures constructed in
Example~\ref{xam:moreinfo}.  As was already observed,
$(\P_3)_* = (\P_4)_*$ (and so $(\P_3)^* = (\P_4)^*$).  
But $(\P_3)_*(b \mid \{b,y\}) = 0$, while
$(\P_4)_*(b \mid \{b,y\}) = 1/2$. 
Thus, even though the upper and lower probability determined by $\P_3$
and $\P_4$ are the same, the upper and lower probabilities determined by 
$\P_3|\{b,y\}$ and $\P_4|\{b,y\}$ are not.
\exam

The following example gives a sense of how conditioning works with sets
of probabilities. 
\xam\label{pch4.xam3} The three-prisoners is the following old puzzle,
which is discussed, for example, by Mosteller \citeyear{Mosteller} and
Gardner \citeyear{Gardner61}:  
\begin{quotation}
\noindent One of three prisoners, $a,$ $b,$ and $c,$ 
has been chosen by a fair lottery to be pardoned, while the other two
will be executed.  Prisoner $a$ does not know who has been pardoned; the
jailer does.  Thus, $a$ says 
to the jailer, 
``Since either $b$ or $c$ is certainly going to be executed,
you will give me no information about my own chances if you give me the
name of one man, either $b$ or $c,$ who is going to be executed.''
Accepting this argument, the jailer truthfully replies, ``$b$ will be
executed.''  Thereupon $a$ feels happier because before the jailer
replied, his own chance of execution was $2/3$, but afterward
there are only
two people, himself and $c,$ who could be the one not executed, and so
his chance of execution is $1/2$.
\end{quotation}

It seems that the jailer did not give $a$ any new
relevant information.
Is $a$ justified in believing that his chances of avoiding execution
have improved?  If so, it seems that $a$ would be equally justified in
believing that his chances of avoiding execution would have improved if
the jailer had said ``$c$ will be executed.''  Thus, it seems that $a$'s
prospects improve no matter what the jailer says!  That does not seem
quite right.

Conditioning
is implicitly being applied here to a
space consisting of three worlds---say $w_a,$ $w_b,$ and $w_c$---where
in world $w_x,$ prisoner $x$ is pardoned.  But this representation of a
world does not take into account what the jailer says.  A better
representation of a possible\index{possible worlds} situation is as
a pair $(x,y),$ where $x,y \in \{a,b,c\}$.  Intuitively, a pair
$(x,y)$ represents a situation where $x$ is pardoned and the jailer says
that $y$ will be executed in response to $a$'s question.  Since the jailer
answers truthfully, $x \ne y$; since the jailer will never
tell $a$ directly that $a$ will be executed, $y \ne a$.
Thus, the set of possible worlds is $\{(a,b), (a,c), (b,c), (c,b)\}$.
The event $\lives \mbox{-}a$---$a$ lives---corresponds
to the set $\{(a,b), (a,c)\}$.
Similarly, the events $\lives \mbox{-}b$
and $\lives \mbox{-}c$ correspond
to the sets $\{(b,c)\}$ and $\{(c,b)\}$, respectively.
By assumption,
each prisoner
is equally likely to
be pardoned, so that each of these three events has probability $1/3$.

The event $says \mbox{-}b$---the jailer says $b$---corresponds
to the set $\{(a,b),(c,b)\}$; the story does not give 
a probability for this event.  
The event
$\{(c,b)\}$ ({\it lives-c}) has probability $1/3$.  But what is the
probability of $\{(a,b)\}$? 
That depends on the jailer's strategy in the one case where he has a choice,
namely, when $a$ lives.  He gets to choose between saying $b$ and $c$ in that
case.  The probability of $(a,b)$ depends on the probability that he
says $b$ if $a$ lives; that is, 
on $\mu(says \mbox{-}b\mid  lives \mbox{-}a)$.

If the jailer 
chooses at random
between saying $b$ and
$c$ if $a$ is pardoned, so that
$\mu(says \mbox{-}b \mid lives \mbox{-}a) = 1/2$,
then $\mu(\{(a,b)\}) = \mu(\{(a,c)\}) = 1/6$,
and $\mu(says \mbox{-}b) = 1/2$.
With this assumption,
$$\mu(lives \mbox{-}a \mid says \mbox{-}b) = \mu(lives \mbox{-}a \inter says \mbox{-}b)/\mu(says \mbox{-}b) =
(1/6)/(1/2) = 1/3.$$
Thus, if $\mu(says \mbox{-}b) = 1/2$,
the jailer's answer does not affect $a$'s probability.

Suppose more generally that $\mu_\alpha,$ $0 \le \alpha \le 1$, is the
probability measure such that $\mu_\alpha(\mbox{{\it lives-}}a) =
\mu_\alpha(\mbox{{\it lives-}}b) = \mu_\alpha(\mbox{{\it lives-}}c) =
1/3$ and $\mu_\alpha(says \mbox{-}b \mid  lives \mbox{-}a)
= \alpha$.  Then straightforward computations show that
$$
\begin{array}{c}\label{prisoners}
\mu_\alpha(\{(a,b)\}) = \mu_\alpha(lives \mbox{-}a) \times
\mu_\alpha(says \mbox{-}b \mid  lives \mbox{-}a) = \alpha/3,\\
\mu_\alpha(says \mbox{-}b) = \mu_\alpha(\{(a,b)\}) +
\mu_\alpha(\{(c,b)\})
=
(\alpha+1)/3,
\mbox{ and}\\
\mu_\alpha(lives \mbox{-}a \mid  says \mbox{-}b) =
\frac{\alpha/3}{(\alpha+1)/3}
= \alpha/(\alpha+1).\end{array}
$$
Thus, $\mu_{1/2} = \mu$.  Moreover, if $\alpha \ne 1/2$
(\ie if the jailer had a particular
preference for answering either $b$ or $c$ when $a$ was the one
pardoned), then $a$'s probability  of being executed would change,
depending on the answer.  
For example, if $\alpha = 0$, then
if $a$ is pardoned, the jailer will definitely say $c$.  Thus, if the
jailer actually says $b,$ then $a$ knows that he is definitely not
pardoned,  that is,
$\mu_0(lives \mbox{-}a \mid  says \mbox{-}b) = 0$.
Similarly, if $\alpha =1$, then $a$ knows
that if either he or $c$ is pardoned, then the jailer will say $b,$
while if $b$ is pardoned the jailer will say $c$.
Given that the jailer says $b,$
from $a$'s point of view the one
pardoned is equally likely to be him or $c$; thus,
$\mu_1(lives \mbox{-}a \mid  says \mbox{-}b) = 1/2$.
In fact,
it is easy to see that if $\P_J = \{\mu_\alpha: \alpha \in [0,1]\}$,
then $(\P_J|says\mbox{-}b)_*(lives \mbox{-}a)
 = 0$ and $(\P_J|says\mbox{-}b)^*(lives \mbox{-}a)  = 1/2$.

To summarize, the intuitive answer---that the jailer's answer gives $a$
no information---is correct if the jailer applies the principle
of indifference in the one case where he has a choice in what to say,
namely, when $a$ is actually the one to live.  If the jailer does
not apply the principle of indifference in this case, then $a$ may gain
information.  On the other hand, if $a$ does not know what strategy the
jailer is using to answer (and is not willing to place a probability on
these strategies), then his prior point probability of $1/3$
\emph{dilates} to the interval $[0,1/2]$.\index{principle of indifference|)}
\exam

As Seidenfeld and Wasserman \citeyear{SeidenfeldW93} have shown,
the dilation phenomenon observed in this example, where the prisoner's
ignorance after hearing the jailer's answer goes from 0---initially
$a$ knew that the probability of him being executed was $1/3$---to $1/2$,
no matter what the jailer says,
is quite general.  Nevertheless, it is easy to see where the dilation is
coming from here, and it is arguably acceptable.  (Although, 
as shown by Gr\"unwald and Halpern
\citeyear{GrunwaldH04}, there may be circumstances when working with
sets of probabilities under which it is most appropriate to ignore new
information and just work with the prior probability.)  
A perhaps more significant problem with this 
approach to conditioning on sets of probabilities 
is that it does not always seem to capture learning, 
as the following example shows.

\xam\label{xam:coins}
Suppose that a coin is tossed twice and the first coin toss is observed
to land heads.  What is the likelihood that the second coin toss lands
heads?  In this situation, the sample space consists of four worlds:\ $hh,$
$ht,$ $th,$ and $tt$.  Let $H^1 = \{hh, ht\}$ be the event that the
first coin toss lands heads.  There are analogous events $H^2,$ $T^1,$
and $T^2$.  Further suppose that all that is known about the
coin is that its bias\index{bias (of a coin)} is either $a$ or $b$,
where $0 \le a < b \le 1$.
The most obvious way to 
represent this seems to be with a set of probability measures $\P =
\{\mu_a, \mu_b\}$.%
\footnote{Some researchers working with probability restrict to sets 
$\P$ of probability measures that are \emph{convex}.  That is, if
$\mu$ and $\mu'$ are both in $\P$, then so is the probability measure
$\alpha \mu + (1-\alpha)\mu'$ for all $\alpha$ in the interval
$[0,1]$ (where $(\alpha \mu + (1-\alpha)\mu')(U) = \alpha \mu(U) +
(1-\alpha)\mu'$; it is easy to check that $\alpha\mu + (1-\alpha)\mu'$
is a probability measure).
I do not make this restriction here, but it is worth noting that nothing
would be lost in this example by taking $\P$ to be the convex set
consisting of all probability meausures $\mu$ such that $a\le \mu(h) \le b$.}
Further suppose that the coin tosses are
independent, so that, in particular, 
$\mu_\alpha(hh) = \mu_\alpha(H^1)
\mu_\alpha(H^2) = \alpha^2$ and that $\mu_\alpha(ht) = \mu_\alpha(H^1)
\mu_\alpha(T^2) = \alpha - \alpha^2$ for $\alpha \in \{a,b\}$.  

Using the definitions, it is immediate that $\P|H^1(H^2) = \{a,b\}
= \P(H^2)$.  At first blush, this seems reasonable.  Since the coin
tosses are independent, observing heads on the first toss does not
affect the likelihood of heads on the second toss; it is either $a$ or
$b$, depending on what the actual bias of the coin is.  However,
intuitively, observing heads on the first toss should also give
information about the coin being used:\ it is more likely to be the coin
with bias $b$.  This point perhaps comes out more clearly if $a= 1/3$,
$b = 2/3$, 
the coin is tossed 100 times, and 66 heads are observed in the first 99
tosses.  What is the probability of heads on the hundredth toss?  
Formally, using the obvious notation, the question now is what 
$\P|(H^1 \inter \ldots \inter H^{99})(H^{100})$ should be.  According to
the definitions, it is again $\{1/3,2/3\}$:~the probability is still
either $1/3$ or $2/3$, depending on the coin used.  But the fact that 66
of 99 tosses landed heads provides extremely strong evidence that the coin
has bias $2/3$ rather than $1/3$.  This evidence should make it 
more likely that the probability that the last coin will land heads is
$2/3$ rather than $1/3$.  The conditioning process does not capture this
evidence at all.  \exam

The inability of this approach to conditioning with sets of
probabilities to capture learning is perhaps its most serious weakness.
Note that this really is a problem confined to sets of
probabilities.   If there is a probability on the possible biases of the
coin, then all these difficulties disappear.  In this case, the sample
space must represent the possible biases of the coin, so there are eight
worlds: $(a,hh), (\beta,hh),
(a,ht), (\beta,ht), \ldots$.  Moreover, if the probability that
it has bias $a$ is $p$ (so that the probability that it has bias
$\beta$ is $1-p$), then the uncertainty is captured by a single
probability measure $\mu$ such that $\mu(a,hh) = pa^2,$
$\mu(\beta,hh) = (1-p)b^2,$ and so on.  A straightforward calculation
shows that $\mu(H^1) = \mu(H^2) = pa + (1-p)b$ 
and $\mu(H^1 \inter H^2) = p a^2 + (1-p) b^2,$ so 
$\mu(H^2\mid H^1) = (pa^2 + (1-p) b^2)/(pa +(1-p)b)$.
With a little calculus, it can be shown that 
$\mu(H^2\mid H^1) = (pa^2 + (1-p) b^2)/(pa +(1-p)b) 
 \ge
\mu(H^2),$ no matter what $a$ and $b$ are, with equality
holding iff $a=0$ or $a=1$.

Intuitively, seeing $H^1$ makes
$H^2$ more likely than it was before, 
despite the fact the coin tosses are independent, because seeing
$H^2$ makes the coin more biased towards heads more likely to be the
actual coin.  This intuition can be formalized in a straightforward way.
Let $C_b$ be the event that the coin has bias $b$ (so that
$C_b$ consists of the four worlds of the form $(b,\ldots)$).
Then $\mu(C_b) = 1-p$ by assumption, while $\mu(C_b\mid H^1) = 
(1-p) b/(p a + (1-p)b) \ge 1-p,$ with equality holding iff $p$ is either
0 or 1 (since otherwise $b/(pa - (1-p)b) > 1$).  
Similarly, if $\mu(H_2 \mid H_1) \ge \mu(H_2)$, with equality holding
iff $p$ is either 0 or 1.

Interestingly, if the bias of the coin is either 0 or 1 (i.e., the coin
is either double-tailed or double-headed, so that $a=0$ and $b=1$), then
the evidence is taken 
into account.  In this case, after seeing heads, $\mu_0$ is
eliminated, so $\P|H^1(H^2) = 1$ (or, more precisely, $\{1\}$), not
$\{0,1\}$.  On the other hand, if the bias is almost 0 or almost 1, say
$.005$ or $.995$, then $\P|H^1(H^2) = \{.005,.995\}$.   Thus, although
the evidence is taken into account in the extreme case, where the
probability of heads is either 0 or 1, it is not taken into account if
the probability of heads is either slightly greater than 0 or slightly less
than 1.

This observation suggests a modification of the conditioning process
that lets us capture learning.  In Example~\ref{xam:coins}, the implicit
assumption is that there is a true bias of the coin, either $a$ or $b$,
which the agent would like to learn.
Given an observation, the \emph{maximum likelihood} approach,
which is standard in statistics, would essentially use the probability
measure that gave the highest probability to the observation from then on.
Since $a < b$ by assumption,
after observing heads, we would use $\mu_b$ for making future
predictions, while after observing tails, we would use $\mu_a$.

The conditioning approach considered so far uses all probability
measures except those that give probability 0 to the observation.
An intermediate approach between these extremes is to consider only
probability distributions that are within some parameter $q$
of
the maximum probability that $U$ gets.
Formally, for $0 < q \le 1$, define 
$$\P^q|U = \{\mu|U: \mu \in \P,  q P^*(U) \le \mu(U)\}.$$
The maximum likelihood approach is a special case of this approach with 
$q=1$.  $\P|U$ as defined earlier, is essentially the case where $q=0$,
except that $\le$ is replaced by $<$.
 
Intuitively, $q$ can be viewed as describing how ``conservative'' the
agent is; the smaller $q$ is, the more conservative the agent.
Note that, for any choice of $q$, learning takes place.  
For example, if we take $\P$ to consist of all the probability measures
$\mu_a$ with $a \in [1/3,2/3]$ (so that the agent considers the
bias of the coin to be somewhere between $1/3$ and $2/3$), and the true
bias is $b \in [1/3,2/3]$, then 
for any choice of $q$ and $\epsilon$, the agent will (with
extremely high probability) converge to considering possible only
distributions $\mu_c$ with $c \in [b-\epsilon,b+\epsilon]$.  
The larger $q$
is, the faster the learning (but the greater the likelihood of making
mistakes by perhaps ignoring a probability measure inappropriately).%
\footnote{Although 
the idea of using a parameter $q$ to do the updating is quite natural,
I have seen it in
print only in the work of Epstein and Schneider \citeyear{r:epstein05a},
who use it in the context of decision making.}

\section{Lower and Upper Expectation}\label{sec:expectation}

In the context of probability and betting games, how much an agent can
expect to win is defined in terms of \emph{expectation}.  

A \emph{gamble} $X$ on $W$ is a function from $W$ to the reals.%
\footnote{A gamble is just a random variable whose range is the reals.}
As is standard in the literature, if $x$ is a real number, take $X=x$
to be the subset of $W$  which $X$ maps to $x$, that is,
$X=x$ is the subset $\{w: X(w) = x\}$.

The \emph{expected value} of $X$ with repect to probability measure
$\mu$, denoted $E_\mu(X)$, is just
$$\sum_x  x \mu(X=x).$$
For example, suppose that the agent bets \$1 and will win \$3 if $U$
happens and lose his dollar if $U$ does not happen.  We can characterize
this bet by the gamble $B = 5X_U - X_{\overline{U}}$, where, for an arbitrary
subset $V$ of $W$, $X_V(w) = 1$ if $w \in V$ and 
$X_V(w) = 0$ if $w \notin U$.   ($X_V$ is called the \emph{indicator
function} for $V$.)

If $\mu(U) = 1/3$, then the agent
expects to win \$5 with probability $1/3$, and to lose \$1 
with probability  $2/3$.  The expected value of this bet is
$$E_\mu(B) = \frac{1}{3} \times 5 + \frac{2}{3} \times (-1) = 1.$$
This seems like an intuitively reasonable characterization of the
agent's expected winnings, provided that his uncertainty is
given by the probability measure $\mu$.

Probabilistic expectation is characterized by some well-known properties.
To make them precise, 
if $X$ and $Y$ are gambles\index{gamble|(} on $W$ and $a$ and
$b$ are real numbers, define the gamble $aX +b Y$ on $W$ in the
obvious way:~$(aX + bY)(w) = aX(w) + bY(w)$.  Say that $X \le Y$ if
$X(w) \le Y(w)$ for all $w \in W$.
Let $\tilde{c}$\glossary{\glosbatilde} denote the
constant function that always returns $c$; that is, $\tilde{c}(w) = c$.

\pro\label{expprop1}
The function $E_\mu$ has the following properties for all 
gambles $X$ and $Y$.
\begin{itemize}
\item[(a)] $E_\mu$ is {\em additive:\/}\index{additivity!for expectation|(}~$E_\mu(X + Y) = E_\mu(X) + E_\mu(Y)$.
\item[(b)] $E_\mu$ is {\em affinely homogeneous:\/}\index{affine
homogeneity|(} $E_\mu(aX + \tilde{b}) = aE_\mu(X) + b$ for all
$a, b \in \IR$.
\item[(c)] $E_\mu$ is {\em monotone:\/}\index{monotonicity, for expectation|(} if $X \le Y,$ then $E_\mu(X)
\le E_\mu(Y)$.
\end{itemize}
\epro

The properties in Proposition~\ref{expprop1} 
essentially characterize probabilistic expectation.

\pro\label{expprop2}
Suppose that $E$ maps gambles on $W$
to $\IR$ and $E$ is additive, affinely homogeneous, and monotone.
Then there is a (necessarily unique) probability measure $\mu$ on $W$ such
that $E = E_\mu$. \epro
Now suppose that uncertainty is represented by a set $\P$ of
probability measures, rather than a single probability measure.
Define
$E_\P(X) = \{E_\mu(X): \mu \in \P\}$.  $E_\P(X)$\glossary{\glosep} is a
set of numbers.
We can use $E_\P$ to define 
obvious analogues of lower and upper probability.
Define the {\em lower expectation\/} and {\em upper
expectation\/}\index{lower/upper expectation|(}
of $X$ with respect to $\P,$ denoted $\lE_\P(X)$\glossary{\glosepl} and
$\uE_\P(X)$\glossary{\glosepu}, as the $\inf$ and $\sup$ of the set
$E_\P(X),$ respectively.   

Just as lower probability determines upper probability (and vice versa),
so lower expectation determines upper expectation.  It is not hard to
show that 
$$\lE_\P(X) = -\uE_\P(-X).$$

We can recover lower and upper probability from lower and upper
expectation.  It is easy to check that $\lE_\P(X_U) = \P_*(U)$ and
$\uE_\P(X_U) = \P^*(U)$, where $X_U$ is the indicator function for $U$
defined earlier.  The converse is not true; lower and upper probability
do not determine lower and upper expectation.

\xam\label{xam:moreinfo1}
Again, consider the sets $\P_3$ and $\P_4$ of probability measures
defined in Example~\ref{xam:moreinfo}.
As observed earlier,
$(\P_3)_* = (\P_4)_*$, and so $(\P_3)^* = (\P_4)^*$.  However, 
if $Y$ is the random variable $X_{\{b\}} - X_{\{y\}}$, then
$\lE_{\P_4}(Y) = \uE(\P_4)(Y) = 0$ (since $\mu(b) = \mu(y)$ for all
probability measures in $\P_4$), while $\lE_{\P_3}(Y) = -1$ and
$\uE_{\P_3}(Y) = 1$.  
\exam

Thus, lower (and upper) expectation can make finer
distinctions than lower and upper probability.  (Note that this is not
the case for probability: $\mu$ determines $E_\mu$ and vice versa.)
Morever, the lower expectation corresponding to a set $\P$ of
probability measures essentially determines $\P$.  

To make this precise, 
recall that a set $\P$ of probability measures on $W$ is \emph{convex}
if, for 
all $\mu, \mu' \in \P$ and $\alpha \in [0,1]$, the probability measure
$\alpha \mu + (1-\alpha)\mu'$ is also in $\P$.  $\P$ is \emph{closed} if
it contains its limits.  That is, 
for all sequences $\mu_1, \mu_2, \ldots$ of probability measures in
$\P,$ if $\mu_n \rightarrow \mu$ in the sense that $\mu_n(U) \rightarrow
\mu(U)$ for all $U \subseteq W,$ then $\mu \in \P$. 
Let $\overline{\P}$ denote the convex closure of $\P$; that is,
$\overline{\P}$ is the smallest closed convex set of probability
measures containing $\P$.  It is easy to see that $\lE_{\P} =
\lE_{\overline{\P}}$ and $\uE_{\P} = \uE_{\overline{\P}}$; adding a convex
combinations of probability measure to $\P$ does not affect the lower
expectation, nor does closing off $\P$ under limits.  The converse
holds as well.

\thm\label{thm:closedconvex} $\lE_{\P_1} = \lE_{\P_2}$ iff
$\overline{\P}_1 = \overline{\P}_2$.  \ethm

Thus, there is a one-to-one map between closed, convex sets of
probability measures and lower expectation functions.  This shows that
lower expectations are essentially as good as sets of probability
measures as representations of uncertainty.  
Walley~\citeyear{Walley91} provides a detailed account of the use of
lower and upper expectations as a representation of uncertainty. 
(He calls them coherent lower and upper \emph{previsions}.)

Lower and upper expectation have a
rather elegant characterization, similar in spirit to (but simpler than)
the characterization of lower and upper probability.  The following
result collects some properties of lower and upper expectation, all of
which are easy to verify.

\pro\label{expprop3}
The functions $\uE_\P$ and $\lE_\P$ have the following properties, for
all gambles $X$ and $Y$.
\begin{itemize}
\item[(a)] $\uE_\P$ is {\em subadditive:}~$\uE_\P(X + Y) \le
\uE_\P(X) + \uE_\P(Y)$;\\
$\lE_\P$ is {\em superadditive:}~$\lE_\P(X + Y) \ge
\lE_\P(X) + \lE_\P(Y)$. \index{subadditivity!for expectation|(}
\index{superadditivity!for expectation|(}
\item[(b)] $\uE_\P$ and $\lE_\P$ are both {\em positively affinely
homogeneous:\/}\index{affine homogeneity!positive|(}
$\uE_\P(a X + \tilde{b}) = a \uE_\P(X) +b$ and $\lE_\P(a X + \tilde{b})
= a \lE_\P(X) +b$ if $a,b \in \IR,$ $a \ge 0$.
\item[(c)] $\uE_\P$ and $\lE_\P$ are monotone.\index{monotonicity, for
expectation|(} 
\item[(d)] $\uE_\P(X) = - \lE_\P(-X)$.
\end{itemize}
\epro

Superadditivity (resp., subadditivity), positive affine
homogeneity, and
monotonicity in fact characterize $\lE_\P$ (resp., $\uE_\P$).

\thm\label{expprop3.5} {\rm \cite{Huber81}}
Suppose that $E$ maps gambles  on $W$
to $\IR$ and is superadditive (resp., subadditive), positively affinely homogeneous,
and monotone.  Then there is a set $\P$ of probability measures on $W$ such
that $E = \lE_\P$ (resp., $E = \uE_\P$).
\ethm

The set $\P$ constructed in Theorem~\ref{expprop3.5} is not unique.
But it follows from Theorem~\ref{thm:closedconvex} that there is a
unique closed convex set $\P$ such that $E = \lE_{\P}$.  $\P$ is actually
the largest set of probability measures $\P'$ such that $E = \lE_{\P'}$,
and consists of all probability measures $\mu$
such that $E_\mu(X) \ge E(X)$ for all gambles $X$.

\section{Decision Making}\label{sec:decision}
One of the standard uses of a representation of uncertainty is to help
make decisions.  Savage \citeyear{Savage} formalizes the decision
process by considering a set $W$ of possible worlds (sometimes called
\emph{states}), a set $C$ of 
\emph{consequences}, and a set $A$ of \emph{acts}, which are functions
from worlds to consequences.  For example, if an agent is trying to
decide how to bet on a horse race, the worlds could
represent the order in which the horses finished the race, and the
consequences could be amounts of money won or lost.  
The consequence of a bet of \$10 on Northern Dancer depends on
how Northern Dancer finishes in the world.  So the bet is an act that
maps worlds (which describe possible orders of finish) to consequences.
The consequence could be
purely monetary (the agent wins \$50 in the worlds where
Northern Dancer wins the race) but could also include feelings (the
agent is dejected if Northern Dancer finishes last, and he also loses
\$10).  

Savage \citeyear{Savage} assumes that the agent has a preference order
$\succeq$ on acts, where $a_1 \succeq a_2$ means that $a_1$ is at least
as good as $a_2$ from the point of view of the agent.  He shows that if
the preference order satisfies certain postulates, 
then the agent is acting as if she has a
probability $\mu$ on worlds, a utility function $u$ mapping consequences
to reals, and is maximizing expected utility; that is, $a_1 \succeq a_2$
iff the expected utility of $a_1$ is at least as high as the expected
utility of $a_2$.  

Savage viewed his postulates as rationality postulates; an agent would
be irrational if her preferences violated the postulates.  However,
as I discussed earlier, in the situation described by
Example~\ref{marbles}, experimental evidence (see \cite{KR95}) shows
that most  people prefer the bet $B_r$ to $B_b$ and also prefers
$B_{by}$ to $B_{ry}$.  These preferences are inconsistent with Savage's
postulates.  Indeed, there does not exist a utility that can be placed on the
two possible consequences (getting \$1 and getting 0) and a probability
that can be placed on $\{b,r,y\}$ such that these preferences correspond
to the order induced by expected utility.

On the other hand,
these preferences can be captured using lower expected utility,
an approach considered by Wald \citeyear{Wald50}, 
G\"{a}rdenfors and Sahlin \citeyear{GS82}, and Gilboa and Schmeidler
\citeyear{GS1989}, among others. 
Taking the obvious set $\P_u$ of probability measures described after
Example~\ref{marbles} and giving utility 1 to winning \$1 and utility 0
to getting 0, it is easy to see that the lower expected utility of act $B_r$
is .3, the lower expected utility of act $B_b$ is 0,
the lower expected utility of
$B_{ry}$ is also .3, and the lower expected utility of $B_{by}$ is .7.
Thus, if the agent prefers the act whose lower expected utility is
larger, then she would indeed prefer $B_r$ to $B_b$ and prefer $B_{by}$
to $B_{ry}$.

Gilboa and Schmeidler \citeyear{GS1989} provide a collection of
postulates that characterize decision making with lower expected utility
in the spirit of Savage's postulates.  Of course, it is debatable
whether these postulates represent ``rationality'' any better than
Savage's do.  However, they do undercut the claim that Savage's
postulate characterize rationality.

Using lower expected utility corresponds to the preference order
$\succeq^1_\P$ on acts  such that 
$\sfa \succeq^1_\P \sfa'$\glossary{\glosaaasucceqpaaa} iff
$\lE_{\P}(u_{\sfa}) \ge \lE_{\P}(u_{\sfa'})$.  
But this is not the only preference rule that can be used if 
uncertainty is represented using a set $\P$ of probabilities.
Other orders can be defined as well:
\begin{itemize}
\item $\sfa \succeq^2_\P \sfa'$ iff $\uE_{\P}(u_{\sfa}) \ge
\uE_{\P}(u_{\sfa'})$;\glossary{\glosaaasucceqpaab} 
\item $\sfa \succeq^3_\P \sfa'$ iff $\lE_{\P}(u_{\sfa}) \ge
\uE_{\P}(u_{\sfa'})$;\glossary{\glosaaasucceqpaac} 
\item $\sfa \succeq^4_\P \sfa'$ iff 
$E_\mu(u_{\sfa}) \ge
E_{\mu}(u_{\sfa'})$ for all $\mu \in \P$.
\end{itemize}

Of course, all of these preference orders reduce to the order provided
by maximizing expected utility if $\P$ is a singleton.  But in general
they are quite different.
The order on acts induced by $\succeq^3_\P$ is very conservative; 
$\sfa \succeq^3_\P \sfa'$ iff the best expected outcome according to
$\sfa$ is no better than the worst expected outcome according to
$\sfa'$.  The order induced by $\succeq^4_\P$ is more refined.
Clearly if
$\sfa \succeq^3_\P \sfa',$ then 
$E_{\mu}(u_\sfa) \ge E_{\mu}(u_{\sfa'})$ for all $\mu \in \P$, so 
$\sfa \succeq^4_\P \sfa'$.
The converse may not hold.  For example,
suppose that $\P = \{\mu, \mu'\}$, and acts $\sfa$ and $\sfa'$ are
such that  $E_\mu(u_\sfa) = 2$,
$E_{\mu'}(u_\sfa) = 4$, $E_\mu(u_{\sfa'}) = 1$, and
$E_{\mu'}(u_{\sfa'}) = 3$.  Then $\lE_\P(u_{\sfa}) = 2$,
$\uE_\P(u_{\sfa}) = 4$,
$\lE_\P(u_{\sfa'}) = 1$, and $\uE_\P(u_{\sfa'}) = 3$, so $\sfa$ and
$\sfa'$ are incomparable according to $\succeq^3_\P,$ yet $\sfa
\succeq^4_\P \sfa'$. 

Which of these rules is the ``right'' one?  We can think of
$\succeq_\P^1$ as representing a very pessimistic agent (who considers
only the worst case); $\succeq_\P^2$ represents an optimistic agent;
while $\succeq^4_\P$ represents an agent who considers all
possibilities.
(I find $\succeq^3_\P$ too conservative, and believe that $\succeq^4_\P$
is a better choice than $\succeq^3_\P$.)  Note that while $\succeq^1_\P$
and $\succeq^2_\P$ place a total order on acts, the ordering
$\succeq^4_\P$ is only partial; some acts will be incomparable under
$\succeq^4_\P$.

\section{Conclusion}
I have provided a brief overview of some of the issues that arise when
representing uncertainty by sets of probabilities, with a particular
focus on updating and decision making.  Before concluding, I briefly
mention two other issues that may be of interest:
\begin{itemize}
\item There are propositional logics for reasoning about about
probability and 
Dempster-Shafer belief functions \cite{FHM}.  More recently, logics
have been provided for reasoning about lower and upper probabilities
\cite{HalPuc00} and lower and upper expectations \cite{HalPuc02:UAI}. 
The syntax of the logics for reasoning about probability, belief
functions, and lower and upper probability are all the same.  All
include statements such as $2/3 l(\phi) + 3/4 l(\psi) \ge 1/2$, where
$\phi$ and $\psi$ are propositional formulas.
The ``l'' here stands for ``likelihood''.  Thus, this statement says 
$2/3$ times the likelihood of $\phi$ plus $3/4$ times the likelihood of
$\psi$ is at least $1/2$.  ``Likelihood'' can be interpreted as either
probability, belief, or lower probability.  In the latter case, 
the upper probability of $\phi$ can be expressed as $1 - l(\neg \phi)$.
(In the case of belief, the same formula defines the plausibility of
$\phi$.)  

The syntax for the logic of expectation is similar in spirit.  It
includes formulas 
of the form $2/3 e(\gamma) + 3/4 e (\gamma') \ge 1/2$, where $\gamma$
and $\gamma'$ are \emph{propositional gambles}.  A propositional gamble
has the form $a_1 \phi_1 + \cdots + a_k \phi_k$, where $a_1, \ldots, a_k$
are real numbers and $\phi_1, \ldots, \phi_k$ are propositional
formulas.  This propositional gamble is interpreted as the gamble
$a_1 X_{\intensionp{\phi_1}} + \cdots + a_k X_{\intensionp{\phi_k}}$,
where $\intensionp{\phi_j}$ is the set of worlds where $\phi$ is true.
Thus, a propositional gamble such as $2 \phi + 3 \psi$ is
interpreted as the gamble $2 X_{\intensionp{\phi}} + 3
X_{\intensionp{\psi}}$, which returns 5 in worlds where both $\phi$ and
$\psi$ are true, 2 in worlds where $\phi \land \neg \psi$ is true, and
so on.   Again, different interpretations of $e$ are allowed; it can be
interpreted as probabilistic expectation, expected belief (see
\cite{Hal31} for a definition of expected belief), or lower
expectation (in which case upper expectation can be defined in the
obvious way).  

The axioms of the logics depend on the interpretation of $l$ and $e$.
In all cases, there is an elegant sound and complete axiomatization.  In
the case of lower and upper probabilities (resp., lower and upper
expectations), not surprisingly, the key axioms are those corresponding
to the properties described in Theorem~\ref{thm:upchar} (resp., 
Theorem~\ref{expprop3.5}).  Moreover, not only are the logics decidable, but
the satisfiability problem is NP-complete in all cases, the same as that of
propositional logic (and of the logic for reasoning about
probability).  Reasoning about lower and upper probability (resp.,
expectation) is thus, in a precise sense, no more difficult than
propositional reasoning. 

\item \emph{Bayesian networks} provide a compact way of
representing probability measures, taking advantage of independencies
and conditional independencies.  There has been a great deal of work in
the AI community showing how Bayesian networks can be used for
efficient probabilistic reasoning (see \cite{Pearl} for an overview).
We can define what it means for $U$ and $V$ to be conditionally
independent with respect to a set $\P$ of probability measures.  Roughly
speaking, $U$ and $V$ are independent with respect to $\P$ if $\mu(V
\mid U) = \mu(V)$ for all $\mu \in \P$ (special care must be taken to
deal with the case that $\mu(U) = 0$; see \cite{Hal25} for details).
Conditional independence is defined in the same way.  Once we do this,
then the whole technology of Bayesian networks can be applied to 
sets of probabilities, essentially without change; see \cite{Hal25} for
details.  

\end{itemize}

As this discussion shows, using sets of probabilities provides a
flexible way of representing uncertainty that enables an agent to
represent ignorance as well as likelihood, while still retaining many of
the pleasant features of using just a single probability measure to
represent uncertainty.

\paragraph{Acknowledgments:}  Thanks to Franz Huber for a careful
reading of the paper and useful comments.
\bibliographystyle{chicagor}
\bibliography{z,joe,riccardo2}

\end{document}